\pdfoutput=1
\documentclass[11pt,final]{article}

\usepackage{acl}
\usepackage{times}
\usepackage{latexsym}
\usepackage[T1]{fontenc}
\usepackage[utf8]{inputenc}
\usepackage{microtype}
\usepackage{inconsolata}
\usepackage{graphicx}

\title{Multilingual European Language Models:\\ Benchmarking Approaches and Challenges}

\author{Fabio Barth \\
  DFKI GmbH, Germany\\
  \texttt{fabio.barth@dfki.de} \And
  Georg Rehm \\
  DFKI GmbH, Germany\\
  Humboldt-Universität zu Berlin, Germany\\
  \texttt{georg.rehm@dfki.de}}

\begin{document}

\maketitle 

\begin{abstract}
The breakthrough of generative large language models (LLMs) that can solve different tasks through chat interaction has led to a significant increase in the use of general benchmarks to assess the quality or performance of these models beyond individual applications. There is also a need for better methods to evaluate and also to compare models due to the ever increasing number of new models published. However, most of the established benchmarks revolve around the English language. This paper analyses the benefits and limitations of current evaluation datasets, focusing on multilingual European benchmarks. We analyse seven multilingual benchmarks and identify four major challenges. Furthermore, we discuss potential solutions to enhance translation quality and mitigate cultural biases, including human-in-the-loop verification and iterative translation ranking. Our analysis highlights the need for culturally aware and rigorously validated benchmarks to assess the reasoning and question-answering capabilities of multilingual LLMs accurately.
\end{abstract}

\section{Introduction}

Recent advances in large language models (LLMs) have demonstrated very strong capabilities across a wide range of natural language processing (NLP) tasks. Their performance is measured and quantified using well-established benchmarks in which the models have to perform tasks such as, among others, reasoning, question answering (QA), or common-sense inference on English text \citep[see, e.\,g.,][]{mmlu, truthfulqa, hellaswag}. The benchmarks consist of text labeled by human annotators \cite{easy2hardbenchstandardized} or they were developed with the help of machine learning techniques that annotate or label the original text data automatically \cite{supergleber}.

Benchmarks are intended to simulate complex problems so that they can provide, when applied to a specific model, an overall impression of how well the model can solve a particular task. Numerous benchmarks exist for a very wide range of tasks, especially with regard to the English language and English-centric LLMs. High-quality multilingual benchmarks for multilingual models are relatively rare. There are non-English monolingual benchmarks developed together with annotators who are native speakers \cite{supergleber} and a few high-quality multilingual benchmarks on reading comprehension, e.\,g., \citet{belebele}. 

Recently, researchers have been using multilingual benchmarks that are comparable across languages \cite{llama2}. For instance, in late 2024, three multilingual LLMs were published that cover all 24 European languages \cite{teuken, euroLLM, salamandra2025}. EuroLLM and Teuken-7B were evaluated using English (MMLU, Hellaswag, TruthfulQA) and a few multilingual benchmarks \cite{mmlu, hellaswag, truthfulqa, belebele}. These evaluations were performed using machine-translated versions of English benchmarks regarding multilingual reasoning, QA, or common-sense inference \cite{multiEUbenchmarks}. Salamandra, however, was evaluated on multilingual datasets that had been human-annotated or human-translated, but do not cover all European languages \cite{baucells-etal-2025-iberobench}.

The translated benchmarks are, like the original English versions, multiple-choice problems, where the model has to choose the correct answer or output from four possible answer options. The correct answer can be calculated by generating the correct output \cite{hellaswag} or measuring the log-likelihood of each possible answer \cite{bear}. The benefit of automatically translated benchmarks is that, besides cost- and time-efficiency in terms of dataset generation, the scores are comparable across languages.\footnote{\url{https://huggingface.co/spaces/occiglot/euro-llm-leaderboard}}

This process has two pitfalls that complicate interpreting the benchmarks' scores. First, the benchmarks are not flawless translations of the English versions. Benchmarks that contain a significant amount of incorrect, incomplete or inadequate translations can distort model predictions -- especially considering that a single incorrectly translated option in a multiple choice setting changes the whole problem space \cite{truthfulqa}. 

To overcome this issue human annotators or an improved and more comprehensive translation pipeline are needed. For instance, a translation-verification process with a human in the loop could help enhance these benchmarks in terms of their translation quality. Another solution could be an iterative process of automatically ranking the translations and re-translating them until the total translation ranking has crossed a certain threshold. There are already models and metrics that can measure and score translation quality \cite{comet}. 

The second problem relates to cultural biases inherently embedded in the English benchmarks that cannot be addressed using automated translations only \cite{globalmmlu}. For example, the MMLU benchmark contains graduate questions from the US that rely heavily on national knowledge (regarding history, religion etc.) \cite{mmlu}. These questions are not meaningful when evaluating the reasoning and QA capabilities of LLMs regarding European languages and cultures.

However, there are already some efforts to translate these benchmarks into multiple languages with regard to cultural biases. Note that these approaches involve human annotators that correct these biases \cite{globalmmlu}.

This paper analyses the most commonly used benchmarks for multilingual LLMs for European languages. We highlight the benefits and interpretability of the existing benchmarks and outline their limitations. 

\section{Approaches}

In this section, we discuss the different approaches to developing multilingual benchmarks. We examine seven multilingual general-domain benchmarks, some of which only partially cover all 24 official EU languages; only one benchmark covers 23 out of 24 languages \cite{belebele}. We analyse benchmarks that include at least nine of the 24 languages. It must be noted that we do not discuss machine translation benchmarks such as FLORES-200, as these assess the quality of translations rather than the performance on general tasks. 

\subsection{Multilingual Benchmarks}
\label{sec:Mulit_bench}

%

The benchmarks can be broadly categorised into two groups based on the way they were developed. On the one hand, multilingual benchmarks are based on English-language benchmarks that were translated into different languages. These benchmarks are either machine-translated or translated by human annotators (occasionally, professional translators). GlobalMMLU \cite{globalmmlu} and MMMLU\footnote{\url{https://huggingface.co/datasets/openai/MMMLU}} belong to this category, they were developed by translating the MMLU benchmark into multiple languages. Two different approaches were used: professional human translators translated MMMLU into 14 languages; using professional human translators is less cost- and time-efficient than simply using automated translation technologies. GlobalMMLU was machine-translated and then annotated by human annotators, organised as an open-source community, using a public annotation framework. As the translation process was automated, \citet{globalmmlu} have been able to translate the original dataset into a total of 42 languages including 11 European languages. To maximise the inclusion of human-translated content, GlobalMMLU also includes human translations of MMMLU. Both data sets show that including human annotators and translators to improve translation quality leads to a smaller task scope and overall sample size \cite{globalmmlu}. 

\citet{multiEUbenchmarks} published EU20, a benchmark that contains translations of five well-established English benchmarks for 20 official EU languages, which cover grade-school science questions \cite{arc}, grade-school math word problems \cite{gsm8k}, common-sense inference \cite{hellaswag}, tests to measure a model's propensity to reproduce falsehoods \cite{truthfulqa}, and elementary mathematics, US history, computer science, law, and multiple choice QA \cite{mmlu}. The benchmarks were translated using the DeepL translation service. The automated translations were not validated by human translators. For the OKAPI benchmark \citet{okapi} translated the same benchmarks as EU20 except for GSM8K. This benchmark differs from EU20 with regard to language coverage and the used translation service. For OKAPI, \citet{okapi} used ChatGPT instead of Deepl. It comprises translations into 26 different languages, including 11 European languages.

The second category of benchmarks includes multilingual datasets that are not based on English data but on regional or multilingual data sources. Here, we can also distinguish between task-specific and multi-task benchmarks. 

Include \cite{include} is a knowledge- and reasoning-centric benchmark across 44 languages, including 15 European languages. It was constructed based on local exam sources in the 44 languages it includes and consists of 197,243 multiple-choice QA pairs. The benchmark was extended with non-English benchmarks like TurkishMMLU \cite{turkishmmlu} or PersianMMLU \cite{persianmmlu}. Include is more extensive than EXAMS \citep{exams}, a cross-lingual and multilingual QA benchmark using high school exams with limited domain coverage.

Belebele \cite{belebele} is a multiple choice machine reading comprehension (MRC) benchmark that covers 122 languages, including all EU languages except Irish. It was developed based on FLORES200 \cite{flores200} and contains a total of 900 questions in a total of 122 languages, each with four multiple-choice answers.

Xtreme \cite{xtreme}, on the other hand, is a comprehensive collection of datasets covering nine tasks and 40 languages (including 15 European languages). As the nine tasks do not cover all 40 languages, an evaluation across the majority of languages is only possible for a subset of tasks \cite{xtreme}. Two datasets (XQuAD and MLQA) have been translated into all 40 languages from the English source using an ``in-house translation tool'' \cite{xtreme}. Overall, Xtreme covers four different categories of tasks: QA, classification, structured prediction, and retrieval.

\subsection{Evaluation Methods}

Most of the benchmarks described in Section~\ref{sec:Mulit_bench} are multiple-choice benchmarks that do not test the multilingual language \emph{generation} but multilingual language \emph{understanding} capabilities of an LLM. Therefore, the metric for evaluating an LLM using one of these benchmarks is accuracy or the harmonic mean of precision and recall (F1 score). The only exception is the benchmark of \citet{xtreme}, which contains classification and structured prediction tasks that are also measured using accuracy and F1-score. However, for the QA tasks of the Xtreme dataset, the exact match metric is also used.

\section{Challenges}

We can identify four main challenges regarding these benchmarks. Below, we explain these and show why these limitations have to be considered when using the benchmarks for the evaluation of LLMs that cover many or all European languages.

\subsection{Challenge 1: Cross-Lingual Comparison}
\label{sec:Cross_Comp}

Besides providing an overview of an LLM's capabilities in various languages, multilingual benchmarks are used to enable cross-lingual comparisons, which highlight a model's capabilities for all languages and show its limits for multilingual use cases. For multilingual benchmarks these comparisons also provide insights into a model's transfer learning capabilities if they have only been trained on a subset of the languages. 

Benchmarks like Xtreme contain a dataset collection covering various tasks, which makes them difficult to interpret for a cross-lingual comparison. Xtreme claims to cover 40 languages; for each task, they provide at least one gold-standard data set. But that does not cover all aspects of each language necessary for transfer because the characteristics of a language can vary depending on the task, domain, and register in which it is used \cite{xtreme}.

To provide such insights, the samples in a benchmark have to be comparable across languages. Benchmarks like GlobalMMLU, Belebele or MMMLU use human annotators and translators to verify the sample comparison \cite{globalmmlu, belebele} while automatically translated benchmarks like \citet{multiEUbenchmarks} or \citet{okapi} are, by definition, unable to enable or guarantee cross-lingual comparisons without a human validation of the translations. Translations can contain tokens or syntactic structures that let the reader backtrack to the original source language. This phenomenon is called translationese \cite{koppel-ordan-2011-translationese}. 
 
\subsection{Challenge 2: Translationese}
\label{sec:Translationese}

Translationese is a linguistic phenomenon characterised by interference from the source language, leading to unnatural structures in the target language \cite{globalmmlu}. In other words, translationese can be described as artifacts or markers in the target translations that can be used to identify the source language \cite{koppel-ordan-2011-translationese}. Typically, these artifacts cause a drop in evaluation quality. On the other hand, models unfamiliar with the source language could benefit from those artifacts in the target translation.

Nevertheless, there are solutions to overcome these linguistic qualms. Professional annotators or translators who validate translations to improve their quality through post-edits can mitigate translationese artifacts in the target language \cite{globalmmlu}. There are also tools for automatic translation quality estimations that use a human-based scoring system to assess the quality of a specific translation \cite{comet, gemba-mqm}. It can be shown that these metrics are sensitive to translationese. The COMET score, for example, varies in terms of its absolute ranking between target translations with and without translationese. However, the results do not show on the system-level ranking \cite{pitfalls_comet}, which averages three lexical segment-level metric scores \cite{comet}. 

\subsection{Challenge 3: Cultural Bias}
\label{sec:Cult_bias}

Most benchmarks evaluate the reasoning capabilities of LLMs. However, reasoning, behavior, and communication are shaped by culture \cite{Tao_2024}. Therefore, cultural biases in multilingual datasets present substantial obstacles with regard to their scope, validity and reliability as global benchmarks \cite{globalmmlu}. Cultural biases will not be mitigated if benchmarks are automatically translated from English into other languages. The published benchmarks apply two different options to address this limitation. First, human annotators evaluate cultural biases present in the original dataset. Global-MMLU, for instance, improved the quality of a multilingual MMLU by engaging with professional and community annotators that label samples as culturally-sensitive or culturally-agnostic \cite{globalmmlu}. Using annotators to verify translations and evaluate cultural bias supports the effectiveness of a multilingual benchmark. 

The second option is to develop a benchmark based on regional resources. The Include benchmark \cite{include} was created based on local exam sources instead of translating benchmarks with inherent cultural bias and debiasing translations. Local exams contain questions about local history, culture, politics, and geographical and regional knowledge.

\subsubsection*{Cultural Sensitivity and Inclusion}

Both approaches described in Section~\ref{sec:Cult_bias} would result in benchmarks with less English-centric cultural biases. However, the first approach mitigates and/or outlines existing biases by identifying whether a dataset is culturally agnostic or not, while the other approach results in benchmarks with cultural biases regarding the intended languages and cultures by not considering English sources that could contain English-centric biases.

The integration of diverse and culturally grounded knowledge is necessary to achieve a certain level of inclusivity and fairness for multilingual evaluations \cite{globalmmlu}. Mitigating cultural bias from benchmarks is fundamental for achieving a certain level of reliability for global benchmarks, but -- as mentioned -- culture shapes reasoning and is, therefore, an essential component in and for communication and for solving multilingual tasks.

\subsection{Challenge 4: Ensuring Data Quality}

Guaranteeing or at least achieving a certain level of data quality is challenging for all benchmarks. The reasons can differ and are also related to challenges already discussed (see Sections~\ref{sec:Cross_Comp} and~\ref{sec:Translationese}). This section will discuss cascading data quality issues that emerge from the source material or the benchmark generation process.

The samples should be evaluated to ensure high data quality in each iteration of the data processing pipeline. This can be done by regular manual or automatic inspections \cite{belebele}. For instance, automatically translated samples can be evaluated using translation scores or human annotators. \citet{multiEUbenchmarks} show that some translations have an error rate of up to 2.3\%, which was identified by manually checking samples with unexpectedly low COMET scores \cite{multiEUbenchmarks}. These error rates differ in individual languages and, therefore, also have an impact on cross-lingual comparability.

As a solution, the scoring and evaluation could be included in the translation process, and lower-rated samples could be re-translated or manually annotated. However, most translated benchmarks rely on manual assessment. \citet{belebele} use manual methods to ensure data quality. Throughout the process of developing the dataset, annotator alignments were defined. Note that forced annotator alignments for translations could increase translationese \cite{10.1162/tacl_a_00317} but it ensures equivalent question difficulty across languages \cite{belebele}.

\section{Conclusions}

In this paper, we analysed, to the best of our knowledge, all benchmarks for the European language area and identified their challenges. The shortcomings of each benchmark are diverse and can be mitigated at different stages of dataset generation. It can also be shown that other approaches for developing datasets can have certain advantages. It can be concluded that a diverse and comprehensive collection of European language benchmarks is beneficial to analyse and compare multilingual language models in detail.

\section*{Acknowledgments}

The work presented in this paper has received funding from the German Federal Ministry for Economic Affairs and Climate Action (BMWK) through the project OpenGPT-X (project no.~68GX21007D).

\bibliography{main}

\end{document}